\documentclass[conference]{IEEEtran}

\IEEEoverridecommandlockouts
\usepackage[utf8]{inputenc}
\usepackage{blindtext}
\usepackage{listings}
\usepackage{adjustbox}
\usepackage{graphicx}
\graphicspath{{figures/draft/}}
\usepackage[hidelinks]{hyperref}
\hypersetup{
  colorlinks   = true, 
  linkcolor={blue!50!black},
  citecolor={blue!50!black},
  urlcolor={blue!50!black}
}

\usepackage{amsmath,amsfonts,amssymb}
\usepackage{lipsum}
\usepackage{placeins}

\usepackage[noadjust]{cite}
 \usepackage{tikz}

\newcommand{\editsA}[1]{#1}
\newcommand{\editsB}[1]{#1}

\usepackage[acronym]{glossaries}
\newacronym{asic}{ASIC}{Application Specific Integrated Circuits}
\newacronym{ai}{AI}{Artificial Intelligence}
\newacronym{cmos}{CMOS}{Complementary Metal Oxide Semiconductor}
\newacronym{iot}{IoT}{Internet of Things}
\newacronym{snn}{SNN}{Spiking Neural Networks}
\newacronym{ic}{IC}{Integrated Circuit}
\newacronym{lif}{LIF}{Leaky Integrate and Fire}
\newacronym{noc}{NoC}{Network on Chip}
\newacronym{qdi}{QDI}{Quasi Delay Insensitive}
\newacronym{bd}{BD}{Bundled Data}
\newacronym{pr}{P\&R}{Place and Route}
\newacronym{aer}{AER}{4 phase handshake Address Event Representation channel}
\newacronym{scnn}{sCNN}{spiking Convolutional Neural Network}
\newacronym{cnn}{CNN}{Convolutional Neural Networks}
\newacronym{soc}{SoC}{System on Chip}
\newacronym{roi}{ROI}{Region Of Interest}
\newacronym{ip}{IP}{Intellectual Property Macro}
\newacronym{fifo}{FIFO}{First In First Out}
\newacronym{sram}{SRAM}{Static Random Access Memory}
\newacronym{tc}{TC}{Temporal Contrast}
\newacronym{dr}{DR}{Dual Rail}
\newacronym{pcfb}{PCFB}{Pre-Charge Full Buffer}
\newacronym{bptt}{BPTT}{Back Propagation Through Time}

\begin{document}

\title{Speck: A Smart event-based Vision Sensor with a low latency 327K Neuron Convolutional Neuronal~Network Processing Pipeline

\thanks{
The Authors want to thank Giacomo Indiveri, Dylan Muir, and Kynan Eng for the creative process of conceiving the design concept.

The Authors want to thank Nicoletta Risi, Madison Cotteret and Hugh Greatorex for their comments, support and advice on the manuscript.
Ole Richter, for the time of the manuscript creation, would like to acknowledge the financial support of the CogniGron research center and the Ubbo Emmius Funds (Univ. of Groningen), during the design and testing Ole Richter was solely affiliated to SynSense AG.\newline
\newline
Affiliations: \\
\textsuperscript{1} SynSense AG, Thurgauerstrasse 60, 8050 Zurich, Swizerland\newline
\textsuperscript{2} SynSense, No. 1577, Tianfu Avenue, Chegdu, Sichuan, PR China \newline
\textsuperscript{3} Bio-Inspired Circuits and Systems (BICS) Lab, Zernike Institute for Advanced Materials, University of Groningen, Netherlands.\newline
\textsuperscript{4} Groningen Cognitive Systems and Materials Center (CogniGron), University of Groningen, Netherlands.\newline

\textsuperscript{*} For inquiries about the publication: o.j.richter@rug.nl, for inquiries about Speck, Samna and Sinabs: sales@synsense.ai and media@synsense.ai
}

}

\author{\IEEEauthorblockA{\textbf{Ole Richter}\textsuperscript{1,3,4,*}, \textbf{Yannan Xing}\textsuperscript{2}, \textbf{Michele De Marchi}\textsuperscript{1}, \textbf{Carsten Nielsen}\textsuperscript{1}, \textbf{Merkourios Katsimpris}\textsuperscript{1}, \\  \textbf{Roberto Cattaneo}\textsuperscript{1},  \textbf{Yudi Ren}\textsuperscript{2}, \editsA{\textbf{Yalun Hu}\textsuperscript{2},} \textbf{Qian Liu}\textsuperscript{1}, \textbf{Sadique Sheik}\textsuperscript{1},    \textbf{Tugba Demirci}\textsuperscript{1,2}, \textbf{Ning Qiao}\textsuperscript{1,2}}}

\maketitle


\begin{abstract}
Edge computing solutions that enable the extraction of high-level information from a variety of sensors is in increasingly high demand. This is due to the increasing number of smart devices that require sensory processing for their application on the edge. To tackle this problem, we present a smart vision sensor \gls*{soc}, featuring an event-based camera and a low-power asynchronous \gls*{scnn} computing architecture embedded on a single chip. By combining both sensor and processing on a single die, we can lower unit production costs significantly. Moreover, the simple end-to-end nature of the \gls*{soc} facilitates small stand-alone applications as well as functioning as an edge node in larger systems. The event-driven nature of the vision sensor delivers high-speed signals in a sparse data stream. This is reflected in the processing pipeline, \editsA{which focuses on optimising highly sparse computation and minimising latency for 9 \gls*{scnn} layers to $3.36\mu s$ for an incoming event}. Overall, this results in an extremely low-latency visual processing pipeline deployed on a small form factor with a low energy budget and sensor cost. We present the asynchronous architecture, the individual blocks, and the \gls*{scnn} processing principle and benchmark against other \gls*{scnn} capable processors.  

\end{abstract}

\section{Introduction}

In order to \editsA{accelerate the development of automated and intelligent systems, an increasing number of data sources will be utilised.} Consequently, the transmission of data between sources needs to be reduced by communicating only relevant and useful information. To achieve this goal, new and radical developments are required in the fields of extreme edge computing and near-sensor processing. The local extraction of relevant information by moving intelligence to the sensory edge poses difficult challenges in real-time processing with low latency and on the smallest of energy budgets. On-demand and sparse computation is a promising solution to reduce computational load and energy consumption~\cite{frenkel_sparsity_2021} but is contrasted with the need for always-on sensory information processing. Event-based processing is a paradigm that can break the trade-off between these two requirements. 

 In the field of image and video processing, \glspl*{cnn} have celebrated significant successes~\cite{redmon_you_2016,krizhevsky_imagenet_2017} and edge inference accelerators for \gls*{cnn}s have been also very successful~\cite{jouppi_-datacenter_2017, reuther_survey_2020}. This is achieved by making \gls*{asic} with more efficient architectures that can skip zero multiplications to sparsify the computational load, and as such are now the industry standard~\cite{albericio_cnvlutin_2016}. To further increase sparsity in the field of event-based computation and sensing, a promising computational method is \gls*{snn}~\cite{basu_spiking_2022, tavanaei_deep_2019}. To exploit highly sparse \glspl*{cnn} even further, event-driven or \glspl*{scnn} only process on the availability of individual pixel data and sparsify the activity from layer to layer by using threshold-based neural units~\cite{sorbaro_optimizing_2020}.

\begin{figure}[!b]
    \centering
    \includegraphics[width=\linewidth]{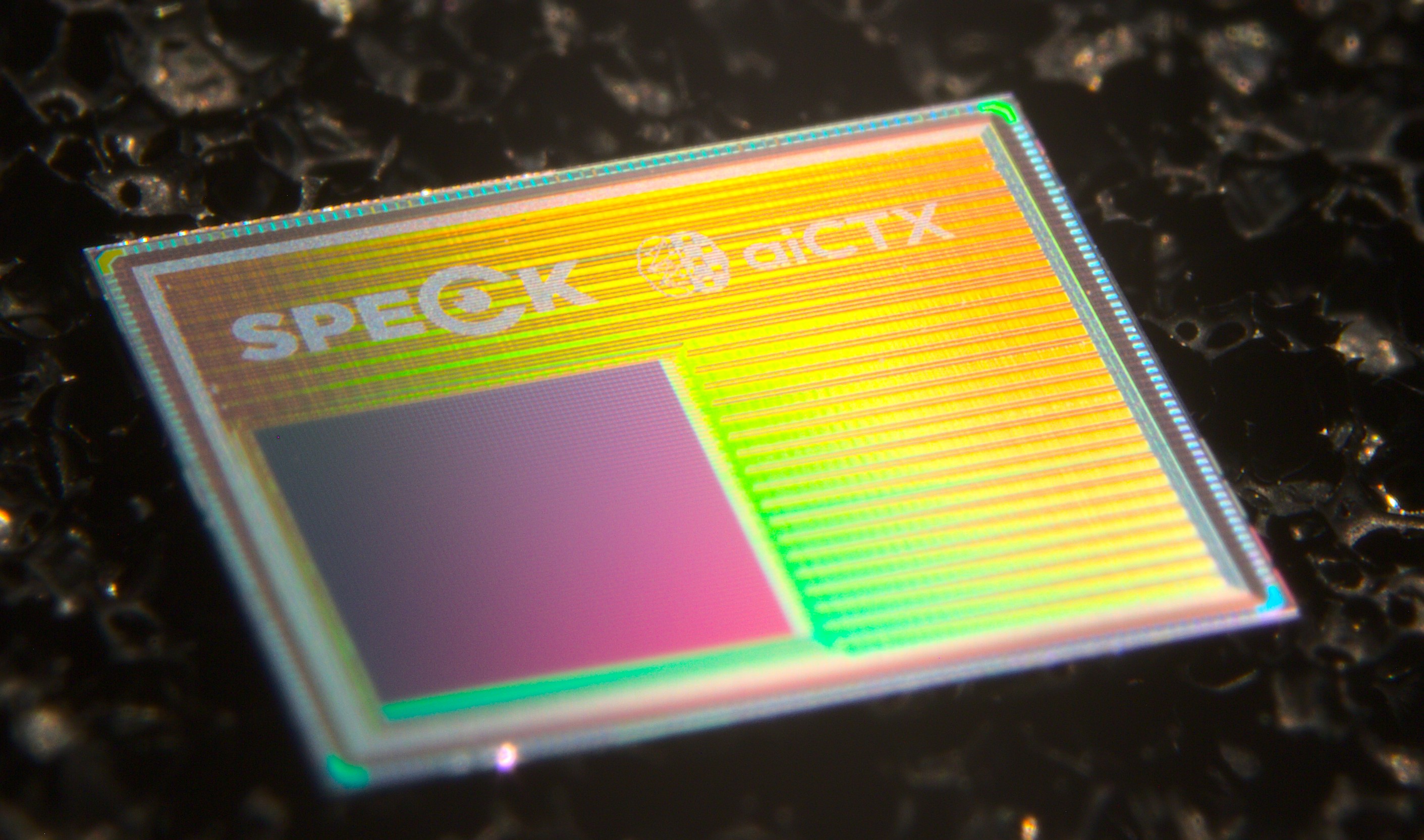}
    \caption{Photograph of the realised "Speck1" \gls*{asic}. The rectangle at the bottom left of the chip is the 128x128 pixel event sensor for machine vision applications, while the rest of the area is occupied by the processing cores and the \gls*{noc}.}
    \label{fig:chip}
\end{figure}
Modern video compression extensively exploits the information redundancy between adjacent frames in order to compress the data stream~\cite{wiegand_overview_2003}. By taking this data reduction approach to the sensor level and enforcing that individual pixel operate independently and communicate only changes in intensity~\cite{lichtsteiner_128x128_2008, posch_qvga_2011, serrano-gotarredona_128x128_2013}, \editsA{the amount of data produced by the sensor is massively reduced. Furthermore, this reduction also reduces the latency of the sensor to levels usually only seen in specialised high frame rate camera modules. The utility of \gls*{tc} encoding cameras has been extensively demonstrated~\cite{gallego_event-based_2022}, and they are now being adopted by industry~\cite{finateu_510_2020}.}

Combining such low latency, high dynamic range and sparse sensor with an event-driven \gls*{scnn} processor~\cite{liu_live_2019}, that excels in real-time low latency processing on a single SoC is a natural technological step. To complement the architectural advantages of always-on sparse sensing and computation, the SoC is built in a fully asynchronous fashion. The asynchronous data flow architecture provides low latency, high throughput processing when requested by sensory input, while immediately shifting to a low power/idle state when the sensory input is absent. Specifically, no complex or slow wake-up procedures must be implemented to reduce power consumption.

Neuromorphic intelligence intends to solve the aforementioned challenges in the following domains: (a) The physical time operation and the processing of always-on sensory signals. The computation speed is matched to natural signals such as 
bio-signals, visual data, speech, gestures, and a wide range of environmental and industrial signals. (b) The redundant information reduction to compute on signal change, sparse data availability, and  statistical and prediction mismatch to considerably enhance power efficiency. (c) Massively parallel computation to keep latency to a necessary minimum. 

Small-scale event-driven \gls*{scnn} processors using different architectural approaches have been proposed in the community~\cite{camunas-mesa_event-driven_2012, frenkel_28-nm_2020, camunas-mesa_configurable_2018, yousefzadeh_fast_2015, zhang_fpga-based_2022, tapiador_morales_neuromorphic_2018, orchard_hfirst_2015}, in contrast to classical large-scale neuromorphic architectures which are able to run \gls*{scnn} networks at a tremendously high synaptic resource cost~\cite{manohar_hardwaresoftware_2022, merolla_million_2014, davies_loihi_2018, schemmel_wafer-scale_2010, painkras_spinnaker_2013, pei_towards_2019}. We present for the first time a resource-efficient medium-scale \gls*{scnn} processor combined with a machine vision event-driven sensor to form a truly novel smart vision sensor on a single \gls*{asic}. 

We first present the asynchronous methodology and principles used in Sec.~\ref{sec:async}., followed by the \gls*{asic} architecture in Sec.~\ref{sec:arch} including the \gls*{scnn} processing principle in Sec.~\ref{sec:conv} and the individual blocks in Sec.~\ref{sec:sensor} to \ref{sec:read}. We conclude with a comparison of different \gls*{scnn} processors and vision sensor processor combinations in Sec.~\ref{sec:discussion}. 

\section{Asynchronous Logic Design Methodology}
\label{sec:async}
\begin{figure}
    \centering
    \includegraphics[width=\linewidth]{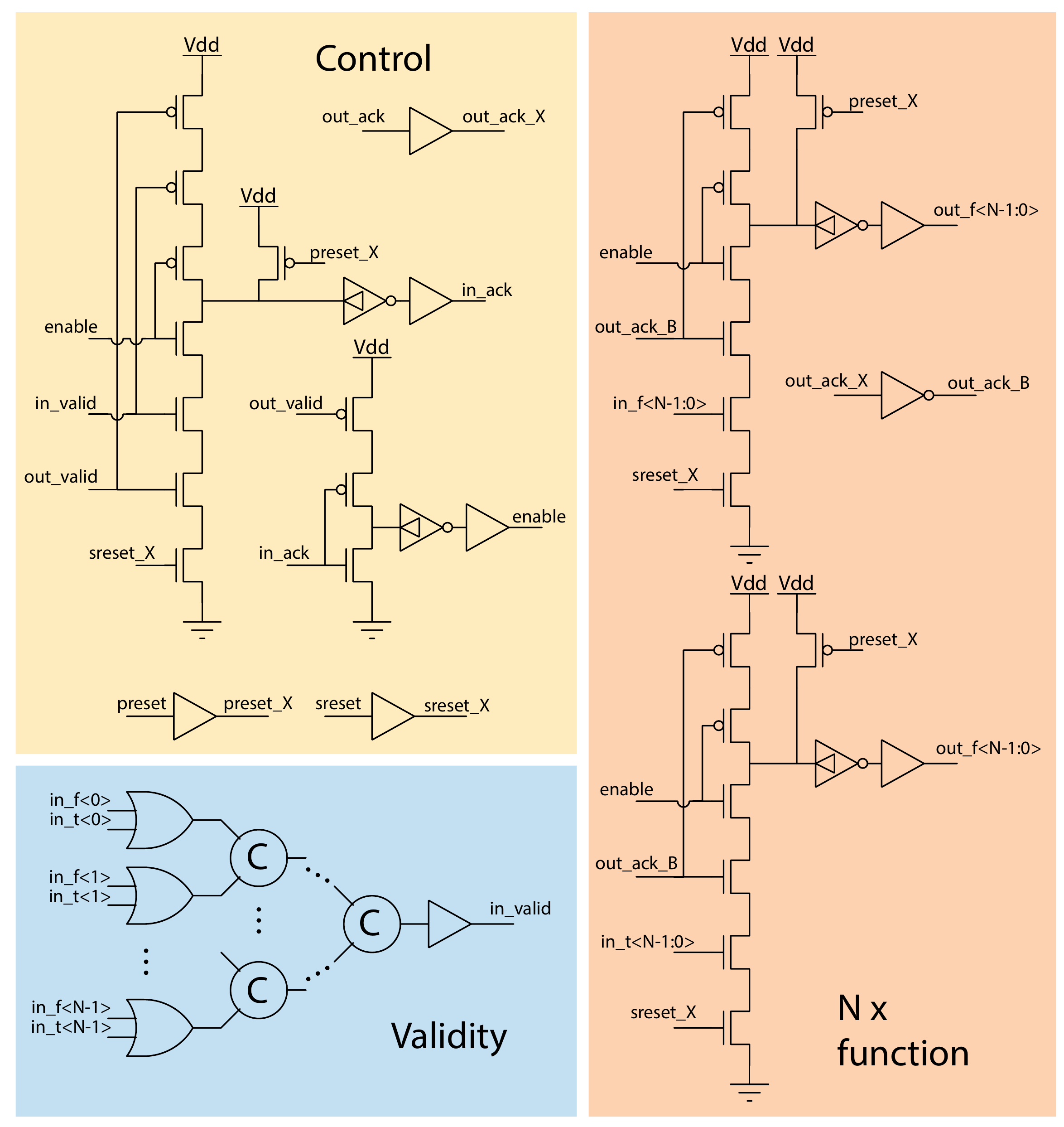}
    \caption{The pull-up/pull-down logic of an N-bit buffer (latch) in Speck1. The control section controls the acknowledge signal of the input channel as well as the transparency of the function block. The function block holds the data and clears it automatically on acknowledgement of the output channel. The Buffer decouples the input channel handshake from the output channel handshake by using the data validity. The validity is a C-element tree with OR2 gates on the input to signal when all data channels have data or all data channels are neutral. To save space, the output validity circuit is shared with the next data flow element. 
 Later versions use a design which supports automated testing.}
    \label{fig:asyncbuffer}
\end{figure}

The design flow of the Speck1 \gls*{asic} followed a golden model approach. The golden model was verified using fully proven applications and ML tasks. Then extensive individual feature tests were derived. These feature tests were run at every step of the \gls*{ic} design flow and compared to the Golden model up to silicon validation after fab-out to verify functionality. Later models have automated production testing support. 
The design methodology follows the dataflow concept~\cite{arvind_dataflow_1986, veen_dataflow_1986, lee_dataflow_1994}, the processing cores are laid out as individual pipeline systems~\cite{sparso_principles_2010, nowick_high-performance_2011}. Processing cores are compositions of template dataflow primitives~\cite{sparso_introduction_2020}: 
Buffer (latch) as seen in Fig \ref{fig:asyncbuffer}, Compose (function block, combined join and/or source), NC-Split (non-conditional fork), Split (conditional demux), Conditional True Pass (conditional sink or latch), Merge (non-deterministic, non-conditional) and Valid-trees (also used as a sink).
The templates are built from pull-up/pull-down state holding logic cells~\cite{martin_programming_1989, van_berkel_beware_1992} and
are derived from \editsA{DYNAP-SE2~\cite{richter_dynap-se2_2024} and DYNAP-SE~\cite{moradi_scalable_2018}. DYNAP-SE was built with the Asynchronous Circuit Toolkit (ACT)~\cite{ataei_open-source_2021} and the template designs follow the \gls*{pcfb} design of~\cite{martin_design_1997,lines_pipelined_1998}. These templates were used due to their low latency design and prior silicon verification available to us.}
In addition, some control primitives such as token latches, forks, muxes, and c-elements that do not carry data, were used to implement data flow control. \editsA{The Speck1 \gls*{asic} uses a 4-phase handshake and the \gls*{qdi} \gls*{dr} data encoding inside the pipelines~\cite{sparso_introduction_2020} of the sensor, sensor event pre-processing, convolutional cores, \gls*{noc} as well as the first half of the readout core pipeline. The encoding is converted to \gls*{bd} encoding for optional off-chip asynchronous event communication as well as for the \gls*{sram} interface, which is using self-timed \gls*{bd} encoding.} The readout core is divided into an asynchronous part, talking to the \gls*{noc}, and a synchronous part to ease integration with standard synchronous \editsA{off-the-shelf components, micro-controllers and infrastructure. 
The combinatorial logic inside the \gls*{qdi} \gls*{dr} pipelines} consists of pull-up/pull-down non-inverting gates, while inversion is modelled by swapping true and false wires of the \gls*{dr} data bits. These non-inverting state-holding gates only model the positive transition of the true or false wire output while the negative signal transition occurs on the reset phase of the handshake. This ensures that the logic is hazard free.
The \gls*{pr} is done by standard industry tools. Performance is ensured by hierarchically detailed automatic floor planning that employs extensive guides and fences for the individual components \editsA{and pipeline stages}.

\section{Architecture}
\label{sec:arch}
\begin{figure}
    \centering
    \includegraphics[width=\linewidth]{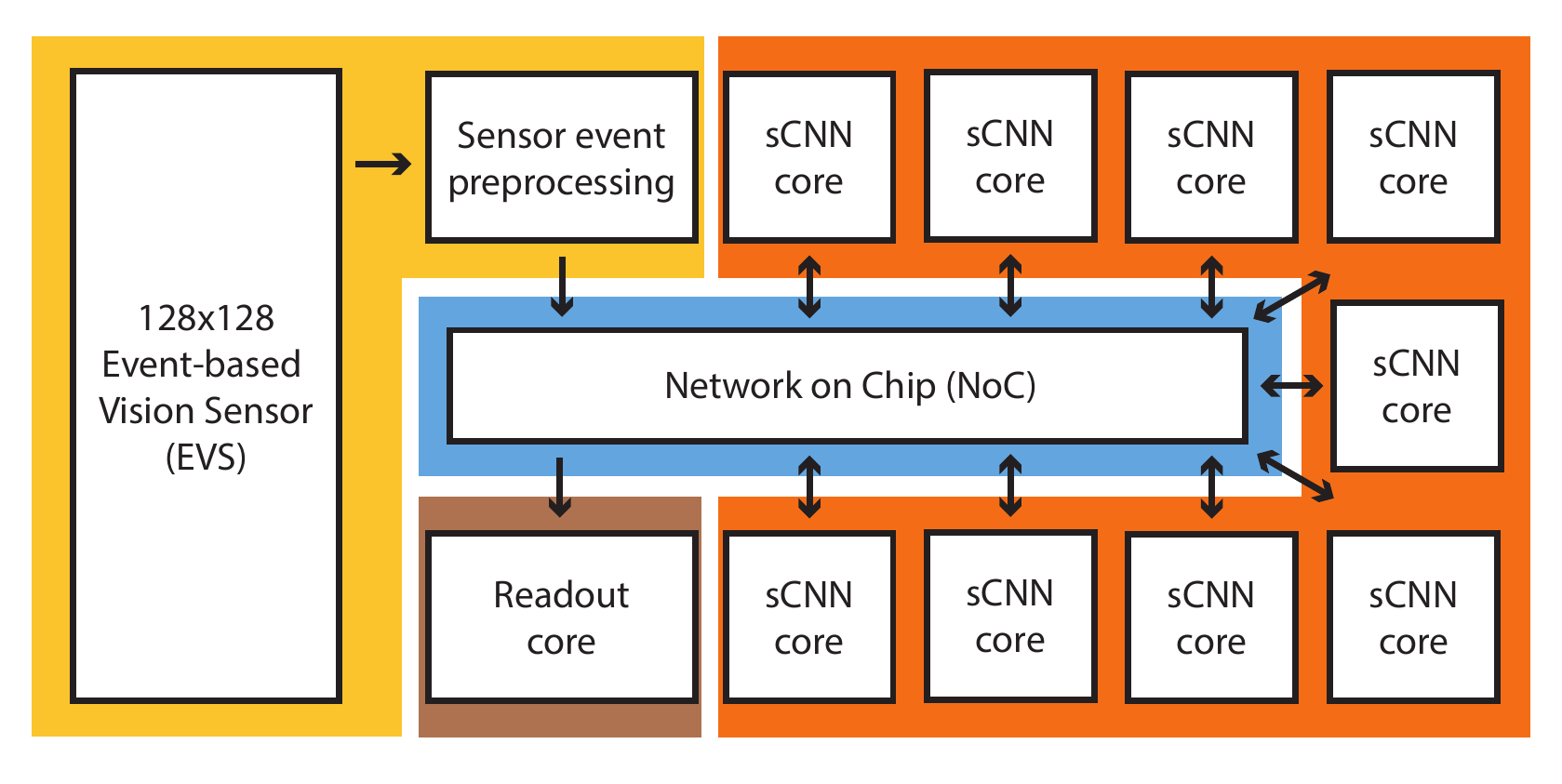}
    \caption{The Speck architecture. The yellow area indicates both the 128x128 event-based vision sensor with its 2D asynchronous readout and the sensor event pre-processing pipeline. The blue area indicates the \gls*{noc} responsible for all the event routing between all the components. The area indicated in orange incorporates all the nine \gls*{scnn} cores that handle one convolution and one pooling layer each. The \gls*{scnn} cores can optionally be operated as fully connected \gls*{snn} layers with some restrictions. The brown area indicates the decision readout logic. This core enables interfacing to simple synchronous periphery. }
    \label{fig:arch}
\end{figure}

The architecture is comprised of 4 different components: the convolution cores, the sensor, the sensor event pre-processing block, and the readout core~\cite{richter_event-driven_nodate, demirci_event-driven_nodate, qiao_data_nodate}. These components are connected by a unicast event routing system the \gls*{noc}, depicted in Fig. \ref{fig:arch}\editsA{.  Walking from the incoming photons through the processing:}

\subsection{Sensor}
\label{sec:sensor}
\begin{figure}
    \centering
    \includegraphics[width=.5\linewidth]{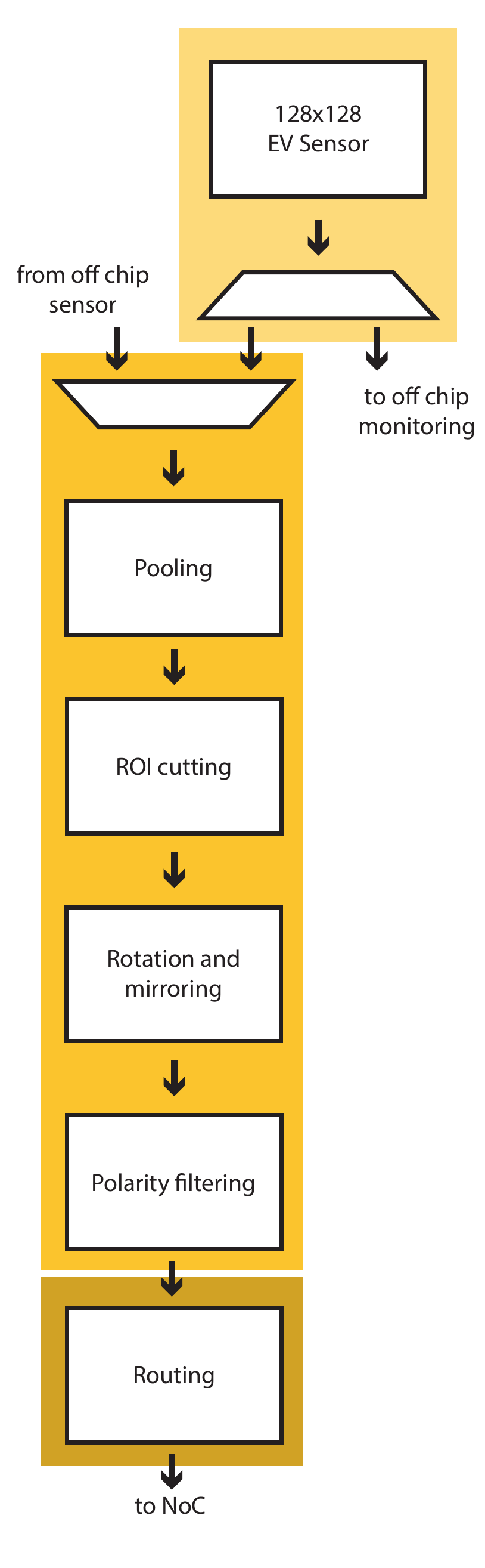}
    \caption{The sensor event pre-processor (Sensor event preprocessing in Fig.~\ref{fig:arch}). The pre-processor can process events from the built-in sensor as well as events from off-chip. The in-built sensor events can also be streamed off-chip for monitoring and further processing. The pipeline stages pool, cut, rotate, mirror, channel filter and shift the input event stream, before forwarding it to one or two destination layers.}
    \label{fig:dvs-pipeline}
\end{figure}
The sensor of Speck1 consists of 128x128 individually operating event-based Vision Pixels, also called Dynamic Vision Pixels~\cite{gallego_event-based_2022}. \editsA{These pixels encode the incident light intensity} temporally on a logarithmic intensity scale, also known as \gls*{tc} encoding. The analog pixel design follows the design given in~\cite{li_132_2019} and was provided by Chenan Li from IniVation AG as an analog pixel \gls*{ip}. Each pixel is attached to a single handshake buffer to decouple the pixel \editsA{reset and timing from the nanosecond delays of the arbitration readout system. From the handshake pixel buffer, the event is handed to the arbitration system by signalling a shared pre-charge pull-down bus in the column and on acknowledge signalling on a shared bus in the row.}
The arbitration is built out of \editsA{one arbiter tree for column arbitration and one for the rows~\cite{purohit_hierarchical_2021, bingham_systematic_2020}. It follows the same design found in DYNAP-SE~\cite{moradi_scalable_2018}. The event address is encoded with a \gls*{qdi} \gls*{dr} encoder from the acknowledge signals of the arbitration trees and handed off as a \gls*{aer} word to the event pre-processing block.}
A complete arbitration process with ID encoding takes approximately $2.5-7.5 ns$ for a single readout. This can be optimised significantly by known techniques~\cite{son_41_2017, fok_serial_2018}, but the specification requirements of real-world signals are met with a margin which allows for a more basic arbitration approach. The arbitration endpoint with buffer \editsA{in the pixel itself is optimised to limit the transistor count and results in a fill factor of $45\%$ front illumination for each pixel. The pixel also has a configurable kill switch to eliminate any hot pixel defects due to fabrication at the pixel level by forcing the pixel and buffer into a reset state.}

\subsection{Sensor event pre-processing core}
\label{sec:pre}

To conform the raw \gls*{aer} event stream from the sensor to the requirements of the \gls*{scnn} a pre-processing stage is required. The image may be flipped, rotated or cropped if only a \gls*{roi} of the image is required. A lower resolution of the image might be required or the polarity can be ignored. To accomplish this, the sensor event pre-processing pipeline consists of multiple stages seen in Fig. \ref{fig:dvs-pipeline}. \editsA{The sensor event will travel through the following pipeline stages:
\begin{itemize}
    \item Sensor interface: the chip can receive pixel events from the built-in sensor and external sources directly via an \gls*{aer} interface and send the sensor events off-chip for monitoring.
    \item Pooling: sum pooling can be used to scale the 2D input address space of every event word by 1:1, 1:2 and 1:4 on $x$ and $y$ coordinate components individually.
    \item \gls*{roi} cutting: cutting can be used to cut a 1x1 to 128x128 size patch out of the 2D input address space that is forwarded to the \gls*{scnn}.
    \item Image rotation and mirroring, in case the smart sensor is mounted sideways, on top or is looking through a mirror, the 2D input address space $x$ and $y$ coordinate components can be flipped, inverted and swapped. 
    \item Polarity filtering: polarity selection enables the selection of both polarities as separate channels, to filter one of them or to combine both polarities on a single channel.
    \item Source mapping: the resulting pre-processed event can be forwarded to up to 2 destination layers via the \gls*{noc}, a routing header is attached and one event is sent per destination.
\end{itemize}
The event is then sent to the convolutional cores via the \gls*{noc}.}
\subsection{Network on Chip}
\editsB{The \gls*{noc} router follows a star topology. The routing system operates in a non-blocking way for any feed-forward network model and routes events via \gls*{aer} connections. The mapping system allows data to be sent from one convolution core to up to 2 other cores and for one core to receive events from multiple sources without addressing superposition with up to 1024 incoming feature channels. On every incoming channel the routing header of every \gls*{aer} packet is read and the payload directed to the destination. This is done by establishing separate physical routing channels that are parallel and do not intersect for any network topology that does not contain recurrence. This prevents skew due to other connections and deadlocks by loops inside the pipeline structure. In combination with the \gls*{pcfb} method the \gls*{fifo} structures display low latency in routing the \gls*{aer} words to their destination. The routing header information is striped from the word during transport, and the payload delivered to its intended destination.}

\label{sec:conv}
\begin{figure}
    \centering
    \includegraphics[width=.48\linewidth]{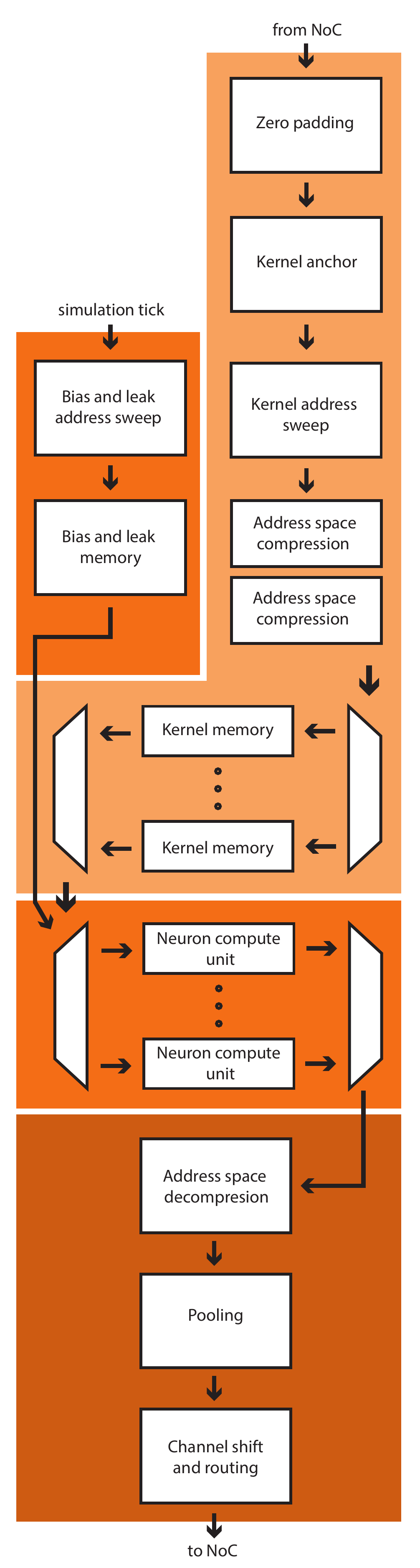}
    \caption{The convolution core architecture (sCNN core in Fig.~\ref{fig:arch}). An event $\{c,x,y\}$ enters the convolution core pipeline, with $c$ as the incoming channel/feature, $x$ as the horizontal coordinate and $y$ as the vertical coordinate. After padding, the event is now expanded to $\{c,x_p,y_p\}$. The Kernel Anchor determines the anchor in both kernel and neuron space $\{c,x_0,y_0,x_0^k,y_0^k\}$. With $x_0,y_0$ being the anchor in the neuron space and $x_0^k,y_0^k$ for the kernel space. The kernel address sweep now calculates the kernel expansion in $x$, $y$ and $f$ the output channel/features to $Z*\{(c,f,x^k,y^k),(f,x,y)\}$, with $Z$ being the synaptic fan-out. The parallel address compression packs the storage addresses compact to avoid unused storage gaps for the neuron $(f,x,y) => n_{\text{comp}}$ and kernel $(c,f,x^k,y^k) => k_{\text{comp}}$. Depending on the core, the kernel memory is split into one or multiple memory blocks for parallel access. The kernel value is read from the storage address $k_{\text{comp}}$, $\{w,n_{\text{comp}}\}$ with $w$ being the signed 8-bit synaptic weight. On a simulation tick, the bias/leak sweep will generate a pair of $\{b_{\text{comp}},n_{\text{comp}}\}$ for every active neuron, the address $b_{\text{comp}}$ gets read in the bias/leak memory and forwarded as $\{w,n_{\text{comp}}\}$ with the kernel events to the neuron. Depending on the core, the neuron unit is split into one or multiple parallel compute units, see Fig.~\ref{fig:neuron_op}. The address space decompression turns the $\{n_{\text{comp}}\}$ back to $\{f,x,y\}$. The sum pooling operates on the same event structure $\{f,x_s,y_s\}$. And the Channel shift and routing prepare it for routing $S*\{d_x,f_s,x_s,y_s\}$ with $S$ being the source fan-out of 1 or 2, $d_x$ corresponding to the destination id and $f_s$ being the arithmetically shifted destination channel.}
    \label{fig:conv_arch}
\end{figure}

\subsection{Convolution Cores}
\begin{figure}
    \centering
    \includegraphics[width=.9\linewidth]{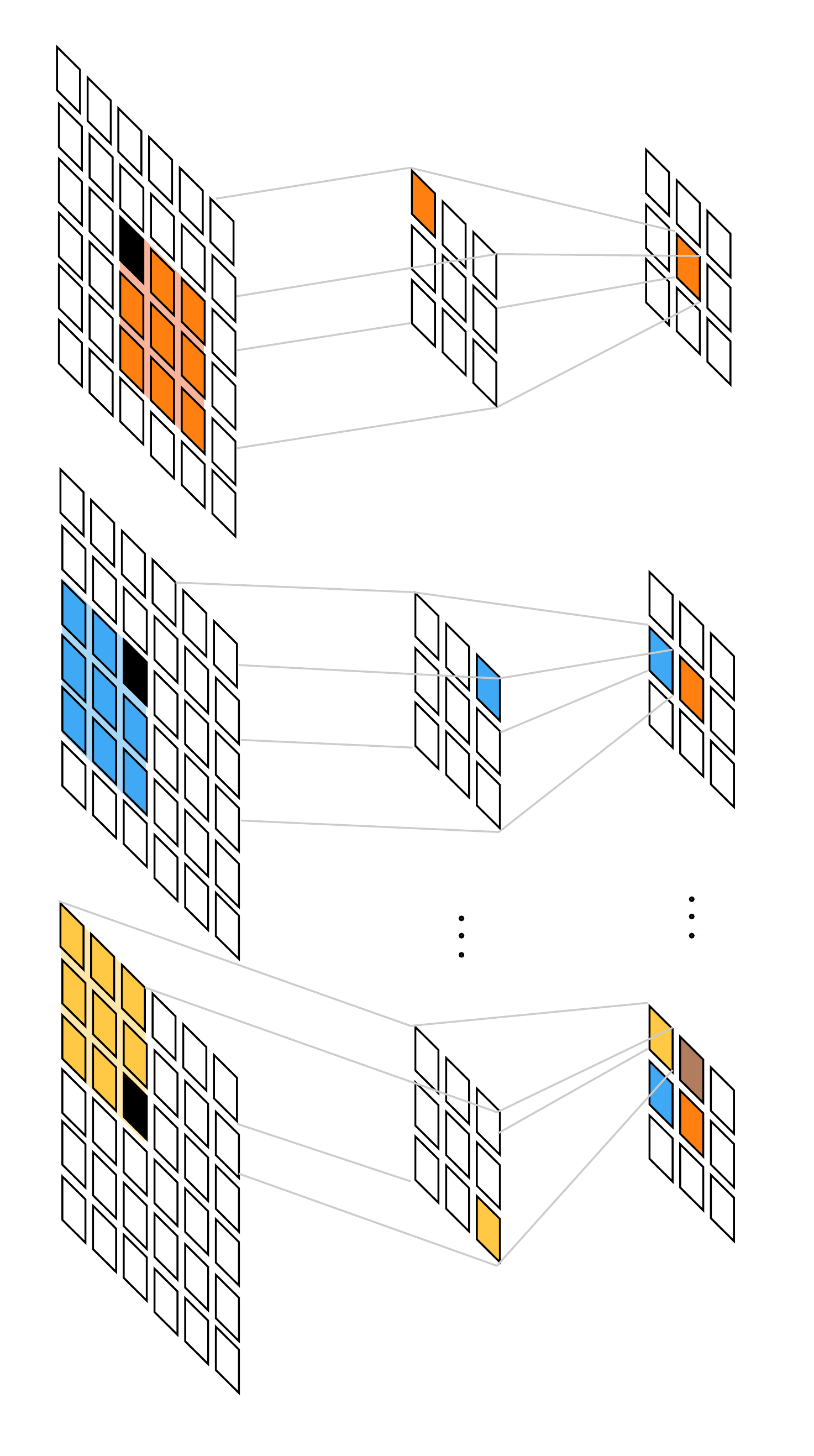}
    \caption{The \gls*{scnn} computation principle. For a single event (black) arriving on the input space, a corresponding anchor coordinate in the kernel space (orange middle) is calculated and the weight at this position is read. \editsA{The read weight is applied to the corresponding anchor neuron in the output space (orange right). The anchor is used to define the starting position in kernel and output space depending on the layer configuration. From this point onward the kernel space is moved from the anchor, in this case with a stride of 1, so by 2 fields in the $x$ coordinate (blue middle), while the neuron is moved on field in the $x$ coordinate from the anchor in the opposite direction (blue right). This is continued until all kernel positions possible have been read. In this case 2 additional - brown and yellow as the stride configuration in this case skips every other position in $x$ and $y$ input coordinate space. This step is repeated for all output channels/features $f$ with their corresponding kernel. The stride and kernel size are configurable and will result together with $x, y, c, f$, with $c$ being the input channel in different sweeps and resulting affected neurons.}}
    \label{fig:scnn_op}
\end{figure}

\begin{figure}
    \centering
    \includegraphics[width=\linewidth]{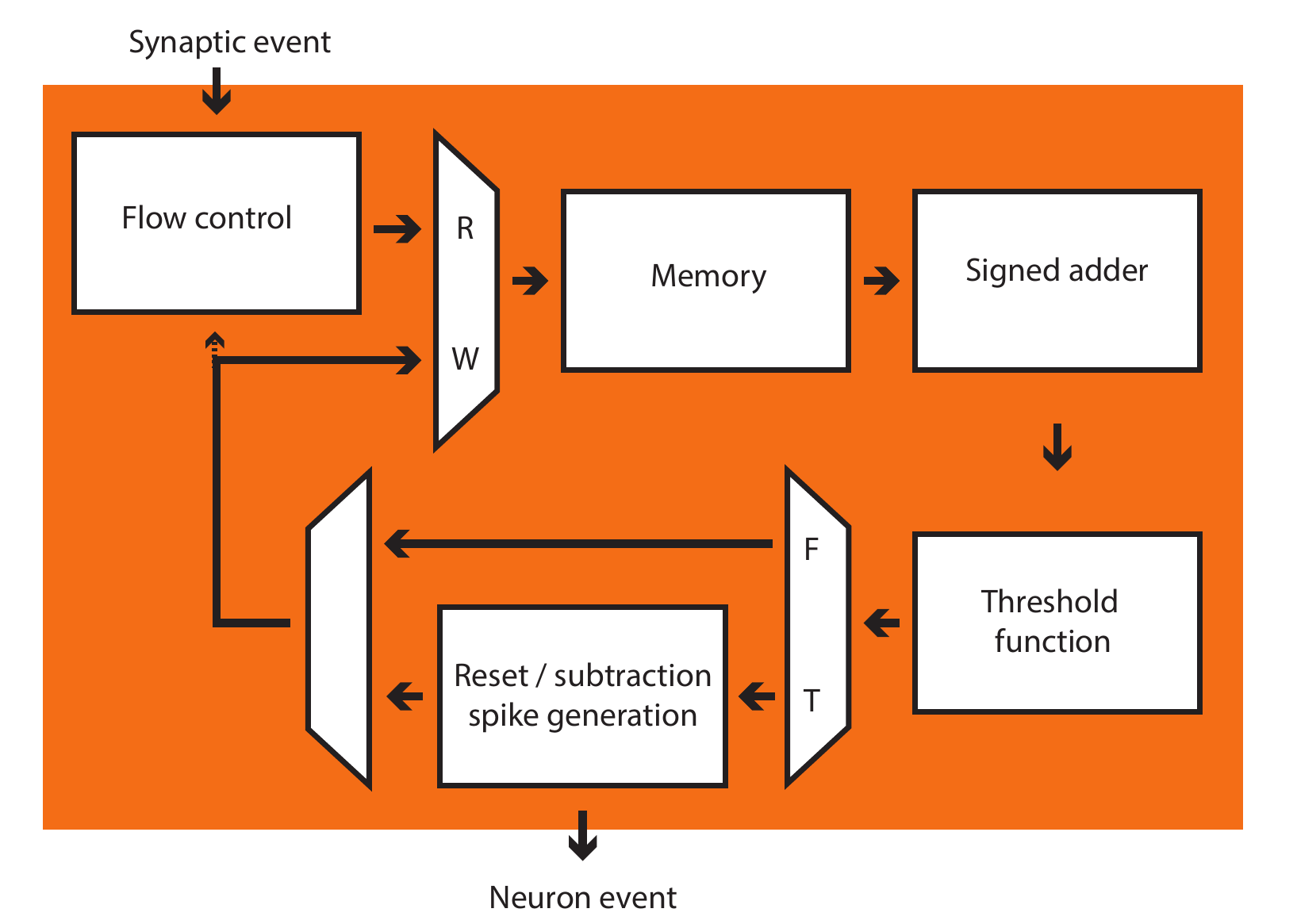}
    \caption{The neuron compute unit from Fig.~\ref{fig:conv_arch}. It uses in-memory-controller compute to model the \gls*{lif} neuron model. The flow control at the input ensures that the controller always has a bubble and is therefore deadlock-free. The signed 16-bit neuron state variable gets read, modified by the signed synaptic or bias input, compared and written back. In case a threshold condition is met, the $\{n_{\text{comp}}\}$ is sent out to indicate the corresponding neuron spiked. The above-threshold condition can both trigger a subtraction operation or a reset to a fixed value of the corresponding neuron state variable. The state variable cannot cross a configured lower bound and will be clamped to that value in case any operation brings the variable below it.}
    \label{fig:neuron_op}
\end{figure}
In contrast to \glspl*{cnn}, event-driven \editsA{\glspl*{scnn} do not operate on a full frame basis: for every arriving pixel event, the convolution is computed for only that pixel position. For a given input pixel, all output }neurons are traversed which are associated with its convolution, as opposed to a kernel that is swept pixel-by-pixel over a complete image. An incoming event includes the $x$ and $y$ coordinates of the active pixel as well as the input channel $c$ it belongs to. 
A step-by-step walk through for an arriving event as seen in Fig~\ref{fig:conv_arch}-\ref{fig:neuron_op}: 
\begin{itemize}
    \item \editsA{Zero padding:} the event is padded to retain the layer size if needed. The image field, i.e. the address of the events, is expanded by adding pixels to the borders to retain the image size after the convolution if needed.
\item \editsA{Kernel anchor and address sweep: In the Kernel mapper, the event is first mapped to an anchor point in the output neuron and the kernel space. The behaviour is described in Fig.~\ref{fig:scnn_op}. Using this anchor the kernel, represented by an address, is linked to an address point in the output space. The referenced kernel is swept over the incoming pixel coordinate. The kernel address and the neuron address are swept inversely to each other as seen in Fig.~\ref{fig:scnn_op}. For every channel in the output neuron space, the kernel anchor address is incremented, so that a new kernel for the new output channel is used. The sweep over the kernel is repeated.  In case a stride is configured in either the horizontal or vertical direction, the horizontal and vertical sweeps are adjusted to jump over kernel positions accordingly. }
\item \editsA{Address space compression: To effectively use the limited memory space, the verbose kernel address as well as the neuron address are compressed to avoid unused memory locations. Depending on the configuration the address space gets packed, so that there are no avoidable gaps inside the address that are not used by the configuration.}
\item Kernel memories: The kernel addresses are then distributed on the parallel kernel memory blocks according to the compressed addresses, and the specific signed 8-bit kernel weight is read. The weight and the compressed neuron address are then directed to the parallel neuron compute-in-memory-controller blocks according to the address location. Kernel positions with 0 weight are skipped during reading and are not forwarded to the neuron.

\item Neuron compute units: The compute-in-memory-controller block model a \gls*{lif} \editsA{neuron with a linear leak} for every signed 16-bit memory word. Besides classic read and write, the memory controller has a read-add-check\_spike-write operation, as shown in Fig~\ref{fig:neuron_op}. Whenever the accumulated value reaches a configured threshold, an event is sent out and the neuron state variable has a threshold subtraction or reset written back. 

\item Bias and leak address sweep and memory: The leak (or bias) is modelled via an additional memory controller. The Leak/bias controller has a neuron individual signed 16-bit weight stored for every output channel map. On a time reference tick an update event with this bias is sent to all its active neurons.  \editsA{The reference tick is supplied from off chip and fully user configurable.}

\item Pooling: The output events are finally merged onto a pooling stage. The pooling stage operates on the sum pooling principle, i.e., it merges the events from 1,2 or 4 neurons in both x and y coordinates individually.

\item Channel shift and routing: Before entering the routing \gls*{noc}, the channels are shifted and a prefix with routing information is  \editsA{added, one event is sent per destination for up to 2.}
\end{itemize}

The individual convolution cores can also be used as fully connected layers with up to 65K,  32K and 16K synaptic connections respectively to model final readout decision layers. \editsA{Fabrication defects in all \gls*{sram} memories for kernel, bias and neuron can be blacklisted with a kill bit per word and are skipped during computation.}

\subsection{Readout core}
\begin{figure}
    \centering
    \includegraphics[width=\linewidth]{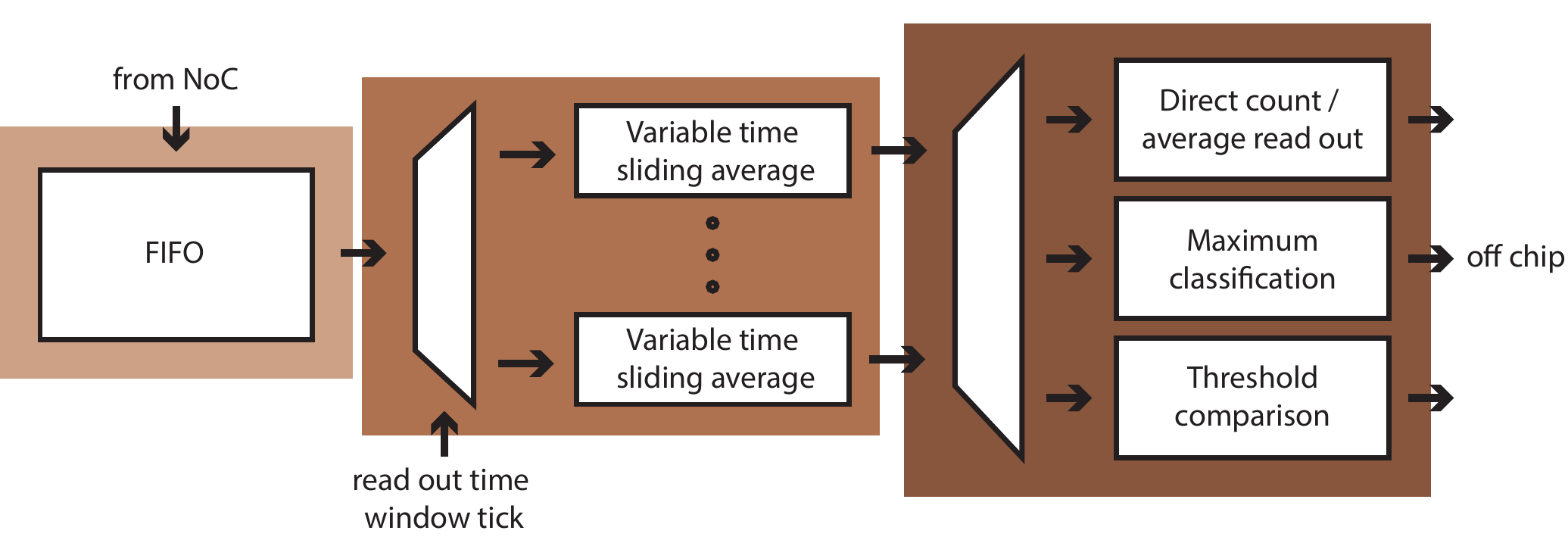}
    \caption{The readout core, from Fig.~\ref{fig:arch}, is separated into an input \gls*{fifo} stage (light brown), a variable time sliding average stage (brown) and a readout processing stage (dark brown). The \gls*{fifo} prevents stalls occurring in the asynchronous to synchronous transition in the following block to affect the \gls*{noc}. The stage that writes to the time sliding average variable is an asynchronous operation whilst in parallel the read in a readout time window is synchronous. The compute stage (dark brown) that does maximum operations and threshold comparisons is fully synchronous to enable easy off-chip interfacing to standard components.}
    \label{fig:readout_Arch}
\end{figure}
\begin{figure*}[t!]
\centering
\begin{adjustbox}{max width=\textwidth}
  \begin{tabular}{l*{8}{c}}
    \hline
     \\
     \textbf{Chip} & \textbf{Speck1} & \textbf{Loihi1} & \textbf{Loihi2} & \textbf{TrueNorth} & \textbf{SCAMP5} & \editsA{\textbf{ISSCC21 Sony}}& \textbf{Spoon} & \textbf{Camunas12} \\
     & [this work] & \cite{davies_loihi_2018,lines_loihi_2018} & \cite{orchard_efficient_2021, intel_technology_nodate} & \cite{merolla_million_2014, akopyan_truenorth_2015} & \cite{carey_100000_2013, liu_-sensor_2022} & \cite{eki_96_2021} & \cite{frenkel_28-nm_2020} & \cite{camunas-mesa_event-driven_2012} \\
     \hline \vspace{1pt} \\
    
    \hline

    Design method  & async & async & async & async & analog+sync & sync & sync & sync \\
    CMOS technology & 65nm & 14nm & 7nm (Intel 4) & 28nm & 180nm & 65nm \& 22nm & 28nm & 350nm \\

    Area incl. I/O & $30mm^2$ & $60mm^2$& $31mm^2$ & $430mm^2$ & $100mm^2$ & $62mm^2$ & $0.32mm^2$& $31.9mm^2$ \\
    Number of neuron & 327.6K & 131K & 1M & 1M & 65K & - & 922 & 4K \\
    Synaptic memory & 272KB & 16MB & 24MB & 32MB & 112KB + analog 
    & 8MB& 750B & 512B \\
    Vision sensor & $128 \times 128$ & No$^2$& No$^1$& No$^2$ & $256 \times 256$ & $4056 \times 3040$ & No$^2$ & No$^1$ \\
    Result readout layer & Yes & Yes$^3$ & Yes$^3$ & No & No & Yes & No & No \\
    
    \hline
    
        Max kernel size  & 16x16 & 64x64 & 64x64$^4$ & 16x16 (1bit) & 4x4  & arbitrary$^4$ & 5x5 & 32x32 \\
            Convolution layers (C) & 9 & arbitrary & arbitrary & arbitrary & sequential & arbitrary$^4$   & 1 & 1 \\
    Pooling layers (P) & 9 & arbitrary & arbitrary &  arbitrary & sequential & arbitrary$^4$  & 1 & - \\
     Features per layer & 1024 & arbitrary & arbitrary & arbitrary & $\frac{\text{65K}}{\text{kernel fields}}$& arbitrary$^4$  & 10 & 32 \\
    
    \hline
    \editsA{8bit syn for 3x3} & 3.01G & 16M$^5$& \textbf{4.29G$^9$} & 9.4M$^{10}$ & - & - & 141K & 1.18M (4bit)$^{11}$ \\
        \editsA{8bit syn for 4x4 }& \textbf{5.35G} & 16M$^5$ & 4.29G$^9$ & 16.7M$^{10}$ & - & - & 250K & 2.1M (4bit)$^{11}$ \\
        \editsA{8bit syn for 7x7 }& \textbf{6.16G} & 16M$^5$ & 4.26G$^9$ & -$^{10}$ & - & - & - & 3.2M (4bit)$^{11}$ \\
    
    \editsA{End-to-end latency} & $1.58\mu s$ (1CP/F)$^8$ & \editsA{$>\Delta t$ (1CP/F)$^6$} & \editsA{$>\Delta t$} (1CP/F)$^6$ & $> 1ms$ (1CP/F)$^6$ & $778\mu s$ (1CP)$^7$ & - & - & \textbf{0.680$\mu s$ (1C)} \\
    \editsA{for sensory event }& \textbf{3.36$\mu s$ (9CP/F)$^8$} & \editsA{$>9\Delta t$ (9CP/F)$^6$} & \editsA{$>9\Delta t$ }(9CP/F)$^6$ & $> 9ms$ (9CP/F)$^6$ & $5595\mu s$ (2CP+2F)$^7$ & \editsA{$4500\mu s$ (28C/P/F)$^7$ }& $117\mu s$ (1CP+2F) & - \\
         \hline
         \\
\end{tabular}
\end{adjustbox}
\centering
    \caption{The table shows the technical specifications of this and related \gls*{asic}s with respect to \gls*{scnn} networks. With the first 4 columns belonging to the group of medium and large systems, the last 2 are small resource-constrained systems built for a limited set of toy applications. The SCAMP5 is an exception as it is an analog cellular processor inside a visual sensor \editsA{and ISSCC21 Sony is a fully integrated standard high resolution image sensor with \gls*{cnn} tensor processor sandwiched together.} Only Speck1 and Loihi2 are specifically built to run larger \gls*{scnn} models. This can be seen in the resulting synaptic counts that are a direct translation of how many different kernels and how often a kernel can be applied. Compared to the general purpose Loihi2 our pure \gls*{scnn} processor archives similar or higher synaptic numbers while having significantly fewer memory resources. As Speck1 is specifically built to consume real-time data, its latency is significantly lower than Loihi1/2 and TrueNorth which architecturally introduce a latency of one timestep \editsA{$\Delta t$ on event generation in each layer.} The small systems are by principle advantages for latency and area as they are limited to running only compact toy networks.\\ \\ Notes: (F) fully connected layer. (1) External sensor interface available. (2) Sensor can be connected with additional hardware. \editsA{(3) Readout interpretation via x86 CPU on \gls*{soc}. (4) Estimated, information not publicly available. (5) Bit packing of 8bit synapse words assumed as described in~\cite{davies_loihi_2018}. Formulae used: }\textit{min(\#neuron per core $\cdot$ \#core $\cdot$ floor(\#inaxon / \#synapses in kernel) $\cdot$ \#syn in kernel, synmem per core $\cdot$ \#core / 8bit)}.  (6) When connected to a sensor and the simulation time is synchronised to real-time, latency per layer is minimum the simulation time resolution \editsA{$\Delta t$, $1ms$ for TrueNorth, as event generation takes one $\Delta t$. (7) full-frame processed. (8) Measured input request to output request edge, when using a 3x3 kernel in every layer, with stride 1, padding 1, pooling has no effect on the latency. (9) For the \gls*{scnn} on-the-fly mapping }no restrictions on kernel sweep in $x$,$y$ and $f$ coordinates assumed: Formulae used: \textit{ \#neuron per core $\cdot$ \#core $\cdot$ floor(\#inaxon / \#synapses in kernel) $\cdot$ \#syn in kernel. \#inaxon} is not mentioned as improved~\cite{orchard_efficient_2021, intel_technology_nodate} and assumed with 4096 as for Loihi1. (10) If \textit{\#synapse in kernel$\cdot$8 $\le$ 256} then \textit{\#neuron per core $\cdot$ \#cores per chip $\cdot$ \#synpses in kernel} else not implementable. (11) Formulae used: \textit{min(lower($\sqrt{\textit{\#kernel memory words}}/\sqrt{\textit{\#synapses in kernel}})^2,\textit{\#max kernel}) \cdot$ \#synapses in kernel $\cdot$ \#neurons }}
    \label{fig:table}
\end{figure*}
\label{sec:read}
The readout core transforms the event data into simple readable values and results. It can simultaneously calculate up to 16 spike class counts or moving averages, where the moving average lengths can be configured and also be set to time bin counting with no averaging. The readout layer optionally compares all values to fixed thresholds, computes the current maximum and makes these results easily accessible on the pins of the chip. All computed values can also be read out for further processing. From the variable time siding average units as seen in Fig.~\ref{fig:readout_Arch} onwards, the circuit performs a clock domain crossing from asynchronous to synchronous. After a time reference clock tick, the previous data is presented and held at the output by standard flip flops and can be interfaced with via standard synchronous processing elements. \editsA{The time reference tick is supplied from off-chip and controlled by the receiving system. }

\section{Results and Discussion}
\label{sec:discussion}
\begin{figure*}
\centering
\begin{adjustbox}{max width=\textwidth}
    \begin{tabular}{lcccccccc}
    \hline \\
    \textbf{Dataset} & \textbf{Chip} & \textbf{Method} & \textbf{Neuron Model} & \textbf{Num Neuron} &\textbf{Num Parameter} & \textbf{On-Chip accuracy}  & \textbf{Power} & \textbf{Energy per inference}  \\
    
    \\
    \hline \vspace{1pt} \\
    
    \hline
     & Loihi1 & SNN TB~\cite{rueckauer_nxtf_2022}& IF & 522 & 597K & 98.43\%  & - & $290\mu J$ \\
     & Loihi1 & SLAYER~\cite{rueckauer_nxtf_2022}& LIF & 522 & 597K & 98.51\%  & - & $620\mu J$\\
    &  Zhang21 & STP~\cite{zhang_28nm_2021} &MLIF& 256 &131K  & 95.7\% & $3.42mW$ & -\\
    NMNIST &  Spoon  & CNN~\cite{frenkel_28-nm_2020} & IF-top1  & 922 & \editsA{750 }& 93.8\% & - & \textbf{0.665}$\mu J$ \\
     & Spoon & CNN-DRTP~\cite{frenkel_28-nm_2020} & IF-top1 & 922 & \editsA{750} & 93\% & - & \textbf{0.665$\mu J$} \\
     & Speck1 & ANN2SNN [this work] &IF& 11K & 9.3K &  86.17\% &  $0.47mW$ & $141\mu J$\\
     & \textbf{Speck1} & \textbf{BPTT-CNN} [this work] & IF &11K& 9.3K & \textbf{98.56\%}  & $0.6 mW$ & $180\mu J$ \\
    \hline \\
    \end{tabular}
    \end{adjustbox}
    \caption{Speck1 performance compared to other architectures. We show that Speck1 matches state-of-the-art performance on \gls*{snn} hardware systems like Loihi1, while our more dedicated approach consumes less energy despite being fabricated in a much older but significantly more cost effective technology. In such a resource friendly benchmark, significantly smaller solutions like Spoon can perform well with some drop in accuracy while gaining a lot on energy efficiency. Most real world applications include more complex or temporal information and require larger networks that are not accessible to small \gls*{scnn} systems. The reported power consumption excludes pad frame power consumption for Speck1. The network structure used for this work is 34x34x2-16C5-16C3-P2-8C3-F10 with 300ms sample exposure, 100 training epochs and a learning rate of 1e-3. 16C5 represents convolutional layer with 5x5 kernel with 16 channels. P2 represents the 2x2 pooling layer and F represents a fully connected layer.}
    \label{tab:results}

\end{figure*}

We propose that a medium-scale pipeline architecture integrated with a sensor can offer improvements both in architecture design and in applicability. \editsA{Matching an event-based sensor with a direct connection to an event based processor avoids incurring delays caused by event batching, as commonly done in event cameras as well as delays due to full frame capture and conversion by industry standard image sensors. Each sensory event is transmitted instantaneously after being encoded by our arbitration system. Our pipeline architecture processes a single event with a latency of $3.36\mu s$ through a nine-layer convolution with pooling network with kernel size $3\times3$ in each layer. This latency was measured by sending an event to and reading the resulting event from the \gls*{asic} with the time between the input and output request edge on the I/O pads. Many events are processed in parallel inside the system as they move step-by-step through the fine-grained pipeline enabling high throughput, set by the neuron compute units, with a measured $\approx30M\mathrm{events/s}$ per unit. Compared to frame-based systems, the event-based nature of the \gls*{soc} can give a classification output as soon as enough evidence is accumulated, as opposed to a frame-based camera with \gls*{cnn} accelerator needing to complete the frame processing in its entirety. As a comparison, the fully integrated, higher resolution ISSCC21 Sony~\cite{eki_96_2021} needs a fixed $10.1ms$ for a $2028 \times 1520$ or $21.3ms$ for a $4056 \times 3040$ full-frame image. To obtain compelling a latency on Speck, the time-to-first-classification (stimulus onset to first classification result) needs to be integrated into the loss function during network training. Looking at classical \gls*{snn} processors, the time} step synchronisation of TrueNorth~\cite{merolla_million_2014} and Loihi1~\cite{davies_loihi_2018}/2 do not allow cores to run controllably out of sync. The latency for each layer is significant when fed with real-time streamed sensory data, as event generation takes one whole simulation time step $\Delta t$ in contrast to our architecture. \editsA{For those systems lowering $\Delta t$ results in a directly proportional trade-off in increase of power consumption }that is not present in this work. The small architectures of Camunas12~\cite{camunas-mesa_event-driven_2012} and Yousefzadeh15~\cite{yousefzadeh_fast_2015} display low latency as they are constrained to small toy networks that don't require a \gls*{soc} and are significantly smaller in specification, reducing power dissipation and transmission delays. SCAMP5~\cite{carey_100000_2013} is another fully-integrated smart vision sensor that follows a very different approach, combining the pixel with simple but powerful analog cellular compute units. They can run simple binary \gls*{cnn}-based networks with quite some overhead~\cite{liu_-sensor_2022}.
\editsA{
A second key point for our presented architecture is the synaptic memory utilisation. Especially for \gls*{cnn} based architectures, the on-the-fly computation of synaptic connections allows for minimising memory requirements. This in turn saves area and energy - in the case of \glspl*{sram}, both dynamic and static. Our dedicated \gls*{scnn} approach allows for many more synaptic connections by using the kernel weights stored in memory and computing all the synapses that share weights compared to standard \gls*{snn} implementations with minimal additional compute required.
In the table in Fig.~\ref{fig:table}, the benefits of a dedicated \gls*{scnn} approach are evident at the 3x3, 4x4 and 7x7 kernel examples where there are magnitudes of differences between the formed synaptic connections. The only other big-scale processor Loihi2~\cite{orchard_efficient_2021} that supports kernel convolution by design restricts its applicable number of formed synapses by its axon-based routing schema with fixed limits on incoming axons per core. For \gls*{scnn}, this network results in either an underutilisation of the many more neurons Loihi2 has compared to our solution or a significant restriction on the in and out feature maps supported. The axon-based routing presents its strength in other network topologies and is widespread in \gls*{snn} processors~\cite{manohar_hardwaresoftware_2022, moradi_scalable_2018, akopyan_truenorth_2015, davies_loihi_2018} which follow a general purpose \gls*{snn} approach.}

\editsA{By offering a simple and small end-to-end system comprising the sensor, the processing, and the decision readout, we offer a unique and easily applicable solution.}
To show the performance of our \gls*{scnn} capability of our \gls*{soc}, we chose a benchmark that can also be run by the smaller systems presented to give a comparison:
The N-MNIST dataset~\cite{orchard_converting_2015} is a spike-converted version of the MNIST dataset~\cite{lecun_gradient-based_1998}. It is recorded using a vibrating ATIS sensor\cite{posch_qvga_2011} with the original images displayed on an LCD monitor. We train a four layer \gls*{scnn} with \texttt{Sinabs}~\cite{sheik_sinabs_2019} using both ANN2SNN~\cite{neil_learning_2016, rueckauer_conversion_2017} and \gls*{bptt} methods. For the training samples, we use the first raw 250ms data out of 300ms and a timestep of 1ms. For training each method, we carried out five repetitions of training with different random initialisations and five repetitions of testing. The testing experiments are performed offline on PC \editsA{and as measurements on Speck1 using a deployed and quantised version of the PC network model.} The events are fed into the \gls*{soc} directly as a raw external sensor event stream bypassing the inbuilt sensor for comparable and repeatable results. For ANN2SNN, we obtained an \gls*{snn} model with an average offline testing accuracy of 94.2{\%} and an on-chip accuracy of 86.17{\%}. For \gls*{bptt} training, it demonstrates state-of-art recognition testing results where offline testing provides an average testing accuracy of {99.3075\%} and on-chip  \editsA{measured testing accuracy of 98.50\%. The samples are presented in real-time so for an inference it takes the length of a sample, with the classification result arriving during the sample presentation as the sample is more comparable to a video. The measured mean of the time-to-first-classification is $18.405ms$ with a standard deviation of $4.818ms$, please note that the network was purely optimised for accuracy and integration of the latency into the loss function can reduce the time-to-first-classification significantly as the hardware itself is optimised for low latency responses.
In Fig.~\ref{tab:results} the results are compared to other \gls*{snn} capable systems. Most prominent are the effects of dedicated \gls*{scnn} hardware systems compared to general purpose \gls*{snn} hardware, so that our proposed system outperforms Loihi1 in accuracy and energy despite the multiple generations older fabrication technology (14nm vs 65nm).} 
\editsB{Additional evaluations on further datasets can be found in \cite{yao_spike-based_2024}. Further applications and demos of Speck with the sensor and \gls*{scnn} combined operation can be seen online. One fall detection demo, were the \gls{soc} is detecting if a human fell and one to detect and follow the human~\cite{lenz_speck_2023}, as well as brief overview of possible application including face and obstacle detection, as well as gesture recognition~\cite{sheik_speck_2022}.}

The training of \gls*{scnn} networks for Speck is supported by the rich, open source, high-level framework \texttt{Sinabs} based on PyTorch and a full development solution called \texttt{Samna}~\cite{nielsen_samna_nodate}. \texttt{Sinabs} can be used for \gls*{scnn} training for this \gls*{asic}. Specifically, it allows optimisations for sparsity and supports the estimation of synaptic operations (SOP) of its networks. 

\section{Conclusion}
We presented a smart sensor fully integrated as a \gls*{soc}, which shifts an efficient \gls*{scnn} architecture directly to the sensor edge. By combining both sensor and processing on a single die into a smart sensor, we lower unit production costs significantly while saving energy on high-speed and low-latency data communication, as the raw sensory data never has to leave the chip. \editsA{The on-the-fly synaptic kernel mapping system lowers the memory resource requirements significantly, making the architecture accessible to larger, more cost effective fabrication technologies. }The event-driven nature of the embedded machine vision sensor delivers high-speed signals in a sparse data manner. The advantages of this combination are taken further by the implemented deep \gls*{scnn} processing pipeline, which is optimised for low latency and \editsA{exploits the benefits of highly sparse computation. This ultimately enables low-latency visual processing on a tiny energy budget for edge and end-to-end applications. Finally, we believe the opportunities and market for different types of smart sensors that follow a similar design principle are very promising.}

\FloatBarrier
\bibliographystyle{IEEEtran}

\bibliography{IEEEabrv,references}

\end{document}